\documentclass{article} 
\usepackage[preprint]{colm2026_conference}

\usepackage{microtype}
\usepackage{hyperref}
\usepackage{url}
\usepackage{booktabs}

\usepackage{lineno}

\usepackage{amsmath}
\usepackage{amssymb}
\usepackage{mathtools}
\usepackage{amsthm}
\usepackage[table]{xcolor}
\usepackage{multirow}
\usepackage{makecell}
\usepackage{fontawesome5}

\definecolor{darkpink}{RGB}{255, 20, 147}
\newcommand{\yes}{\checkmark}

\definecolor{darkblue}{rgb}{0, 0, 0.5}
\hypersetup{colorlinks=true, citecolor=darkblue, linkcolor=darkblue, urlcolor=darkblue}

\title{DARE: Diffusion Large Language Models Alignment and Reinforcement Executor}

\author{
Jingyi Yang\thanks{Equal contribution} ~ \&
Yuxian Jiang\footnotemark[1] ~ \&
XuHao Hu\footnotemark[1] \\
Shanghai Artificial Intelligence Laboratory\\
Fudan University\\
\texttt{yangjingyi946@gmail.com} \\
\And
Shuang Cheng  \\
Shanghai Artificial Intelligence Laboratory\\
Zhejiang University\\
\And
Biqing Qi \& Jing Shao\thanks{Corresponding author} \\
Shanghai AI Laboratory
}

\begin{document}

\ifcolmsubmission
\linenumbers
\fi

\maketitle

\qquad \qquad \qquad \qquad \quad \faGithub \, \textbf{Code:} \href{https://github.com/yjyddq/DARE}{\textcolor{darkpink}{https://github.com/yjyddq/DARE}} \\

\begin{abstract}
Diffusion large language models (dLLMs) are emerging as a compelling alternative to dominant autoregressive models, replacing strictly sequential token generation with iterative denoising and parallel generation dynamics.
However, their open-source ecosystem remains fragmented across model families and, in particular, across post-training pipelines, where reinforcement learning objectives, rollout implementations and evaluation scripts are often released as paper-specific codebases. 
This fragmentation slows research iteration, raises the engineering burden of reproduction, and makes fair comparison across algorithms difficult.
We present \textbf{DARE} (\textbf{d}LLMs \textbf{A}lignment and \textbf{R}einforcement \textbf{E}xecutor), an open framework for post-training and evaluating dLLMs. Built on top of verl~\cite{sheng2024hybridflow} and OpenCompass~\cite{2023opencompass}, DARE unifies supervised fine-tuning, parameter-efficient fine-tuning, preference optimization, and dLLM-specific reinforcement learning under a shared execution stack for both masked and block diffusion language models. 
Across representative model families including LLaDA, Dream, SDAR, and LLaDA2.x, DARE provides broad algorithmic coverage, reproducible benchmark evaluation, and practical acceleration. 
Extensive empirical results position that DARE serves as a reusable research substrate for developing, comparing, and deploying post-training methods for current and emerging dLLMs. 
\end{abstract}

\section{Introduction}
\label{sec:introduction}

Diffusion large language models (dLLMs)~\cite{nie2025large,ye2025dream,cheng2025sdar,khanna2025mercury,geminidiffusion,song2025seed} have rapidly evolved from an intriguing alternative to autoregressive language modeling into a growing family of practical model architectures.
Early masked diffusion large language models such as LLaDA~\cite{nie2025large} and Dream~\cite{ye2025dream}, as well as recent block diffusion or semi-autoregressive variants such as SDAR~\cite{cheng2025sdar} and LLaDA2.x~\cite{bie2025llada2,bie2026llada2.1}, show that diffusion-style generation can support flexible token order, bidirectional conditioning, and parallelism.
As these models mature, however, the main bottleneck is shifting away from model definition and toward post-training and evaluation infrastructure.

The current open-source dLLM ecosystem is highly fragmented.
Most dLLM-based reinforcement learning (RL) methods are released as paper-specific repositories, each with its own model fork, rollout implementation, reward interface, and evaluation scripts.
This fragmentation creates at least three problems.
First, it slows research iteration because integrating a new model or objective may requires re-implementing the surrounding infrastructure.
Second, it makes cross-paper comparison unreliable because algorithmic differences are entangled with execution and evaluation differences.
Third, it raises the engineering barrier for researchers who want to study, use and extend prior work. 
This systems gap is becoming more important, not less.
Importantly, it cannot be solved by directly reusing LLM RL frameworks.
Most LLM post-training pipelines assume left-to-right generation, sequence log-probabilities over a single decoding trajectory, and rollout engines built for autoregressive serving.
dLLMs instead require diffusion-aware forward and reverse processes, denoising-state likelihood surrogates, and model-family-specific rollout backends.

To address this need, we present \textbf{DARE} (\textbf{d}LLMs \textbf{A}lignment and \textbf{R}einforcement \textbf{E}xecutor), a unified post-training and evaluation framework for diffusion large language models.
DARE is built on top of verl~\cite{sheng2024hybridflow} for training and OpenCompass~\cite{2023opencompass} for evaluation, while adding the missing dLLM-specific execution layers or modules.
It integrates a broad set of post-training recipes, including supervised fine-tuning (SFT), parameter-efficient fine-tuning (PEFT), preference optimization, and multiple dLLM-specific RL algorithms, and exposes them through a shared pipeline.
The framework supports both masked diffusion language models (MDLMs) and block diffusion language models (BDLMs), enabling one infrastructure stack to serve LLaDA~\cite{nie2025large}, Dream~\cite{ye2025dream}, SDAR~\cite{cheng2025sdar}, LLaDA-MoE~\cite{zhu2025llada-moe}, and LLaDA2.x~\cite{bie2025llada2,bie2026llada2.1} families.

DARE also treats systems optimization as a first-class part of dLLM post-training.
For MDLMs, the framework decouples the attention backend used during rollout from the one used during actor updates: rollout uses Fast-dLLM~\cite{wu2025fast} together with FlashAttention backends~\cite{dao2022flashattention,dao2023flashattention}, while training uses variable-length backends of FlashAttention to reduce padding overhead.
For BDLMs, DARE integrates LMDeploy~\cite{2023lmdeploy,zhang2025efficient}, SGLang~\cite{sglang2024}, and fused loss kernels for acceleration, and integrates FlexAttention~\cite{dong2024flex} for training.
In addition, DARE extends OpenCompass~\cite{2023opencompass} with dLLM-aware evaluation so that benchmarking is part of the same reusable framework.
DARE is not proposed at the method-level. Instead, it is a reusable research substrate that makes it possible to integrate, compare, and evaluate rapidly growing dLLM post-training methods within a unified execution environment.

Our contributions are four-fold:
\begin{itemize}
    \item We present DARE, a unified framework that brings together dLLM post-training and evaluation under a single open-source stack, spanning masked and block diffusion language model families.
    \item We integrate a broad range of post-training methods, including SFT, PEFT, preference optimization, and multiple dLLM-tailored RL algorithms, enabling fairer algorithm comparison.
    \item We implement extensive dLLM-specific system optimizations for rollout, training, and evaluation.
    \item We empirically show that DARE provides broad model and benchmark coverage while turning fragmented paper-specific implementations into a more reproducible and comparable dLLM research workflow.
\end{itemize}

\section{Related Works}
\label{sec:related works}

\subsection{Diffusion Language Models}
\label{subsec:diffusion language models}

\textbf{Masked Diffusion Language Models.}
D3PM~\cite{austin2021structured} formulates the discrete diffusion by modeling a sequence of corruption processes over categorical variables, for example with an absorbing state ($\texttt{<MASK>}$) or uniform noise~\cite{austin2021structured,wu2023ar,sahoo2024simple,lou2023discrete,zheng2024masked,gong2024scaling,ou2024your,nie2024scaling}. 
For masked diffusion language models, the categorical distribution $q(x_t^\ell|x_0^\ell)$ is parameterized by a linear interpolation between the original one-hot vector $\mathbf{x}_0^\ell$ and the absorbing vector $\mathbf{m}$ ($\texttt{<MASK>}$ token):
\begin{equation}
\label{eq:forward process}
q(x_t^\ell|x_0^\ell) = \text{Cat}(x_t^\ell; \alpha_t\mathbf{x}_0^\ell + (1-\alpha_t)\mathbf{m}),
\end{equation}
where $\alpha_t$ denotes the noise schedule at timestamp $t$. The mask predictor $p_{\theta}$ is trained to reverse the corruption process by minimizing the negative evidence lower bound objective (NELBO):
\begin{align}
\label{eq:mdlm_loss}
\mathcal{L}_\theta = & \mathbb{E}_{x_0 \sim p_{\text{data}}, {x}_t \sim q({x}_t|{x}_0), t\sim \mathcal{U}(0,1]} \left[-\frac{1}{t} \sum_{\ell=1}^{L} \mathbb{I}[{x}_t^\ell = \texttt{<MASK>}] \log p_{\theta}({x}_0^\ell | {x}_t) \right].
\end{align}
Recent work~\cite{nie2025large,ye2025dream,khanna2025mercury,geminidiffusion,song2025seed} has scaled this paradigm to large language modeling.
These models motivate diffusion-specific rollout and training pipelines because their generation process differs fundamentally from autoregressive left-to-right decoding.

\textbf{Block Diffusion Language Models.}
In parallel, block diffusion paradigms~\cite{han2022ssd,arriola2025block,fathi2025unifying,bie2025llada2,bie2026llada2.1} have produced models such as SDAR~\cite{cheng2025sdar} and LLaDA2.x~\cite{bie2025llada2,bie2026llada2.1}, which combine intra-block diffusion with inter-block autoregression and support variable-length generation with kv-caching.
Specifically, a sequence $x$ is partitioned into $B$ contiguous, non-overlapping blocks, $\{x^1, \dots, x^B\}$, each containing $L'=\frac{L}{B}$ tokens. The likelihood factorizes over block as $\log p_{\theta}(x) = \sum_{b=1}^{B} \log p_{\theta}(x^{b} | x^{<b})$, and block-wise NELBO is given by:
\begin{align}
\label{eq:bdlm_loss}
\mathcal{L}_\theta = & \mathbb{E}_{x_0 \sim p_{\text{data}}, b \sim \mathcal{U}[1, B], t\sim \mathcal{U}(0,1]} \left[-\frac{1}{t} \sum_{\ell=1}^{L'} \mathbb{I}[{x}_t^{b,\ell} = \texttt{<MASK>}] \log p_{\theta}({x}_0^{b,\ell} | {x}_t^b, x^{<b}) \right]
\end{align}
where ${x}_0^{b,\ell}$ and ${x}_t^b$ denote the clean and corrupted sequences of block $b$, respectively. The $p_{\theta}(x^{b} | x^{<b})$ is trained to recover the clean block $x^b$ from its noisy counterpart conditioned on preceding clean blocks $x^{<b}$.
This regime introduces different systems requirements from masked diffusion, especially for rollout backends, block-wise verification, and online policy updates. The existence of both MDLM and BDLM families is one reason a unified dLLM executor is needed.

\subsection{Post-Training Frameworks and Reinforcement Learning for dLLMs}
\label{subsec:post-training frameworks and reinforcement learning for dLLMs}

Since 2025, dLLM post-training has developed rapidly.
Early work has adapted policy gradient-based and RL algorithms to masked diffusion large language models (MDLMs). For example, VRPO~\cite{zhu2025llada} is designed for human preference alignment in dLLM. d1~\cite{zhao2025d1} and Coupled-GRPO~\cite{gong2025diffucoder} directly adapt GRPO~\cite{chen2025conditional,shao2024deepseekmath,guo2025deepseek} for one-step denoising optimization. MDPO~\cite{he2025mdpo} addressed the training-inference mismatch by optimizing the progressive refinement schedule used at inference. CJ-GRPO~\cite{yang2025taming} emphasized consistency between rollout and optimization trajectories, while DiFFPO~\cite{zhao2025diffpo} studied off-policy surrogate policies and joint optimization of reasoning quality and inference-time efficiency. In addition, SPG~\cite{wang2025spg} and BGPO~\cite{lin2025boundary} introduce alternative upper-lower bounds beyond ELBO for stable optimization.
However, MDLMs may face exploration limitations due to fixed-length generation. Variable-length generation strategies~\cite{li2025beyond,yang2026rho} are expected to be integrated to unlock this constraint. 
In contrast, the block diffusion paradigm can adaptively determine the generation length, and their corresponding RL algorithms (e.g., TraceRL~\cite{wang2025revolutionizing}, DiRL~\cite{zhu2025dirl} and EBPO~\cite{bie2026llada2.1}) account for semi-autoregressive properties, explicitly emphasizing the order of rollout trajectories.
These developments also illustrate why LLM RL frameworks cannot be transferred to dLLMs directly.
LLMs typically employ next-token-prediction, equipped with exact sequence log-probabilities and inference engines, whereas dLLMs optimize over denoising trajectories, ELBO-style or diffusion-aware surrogates, and model-specific forward/reverse processes.

\section{DARE Framework}
\label{sec:dare framework}

\subsection{Overview}
\label{subsec:overview}
\begin{figure*}[t!]
\centering
\includegraphics[width=\textwidth]{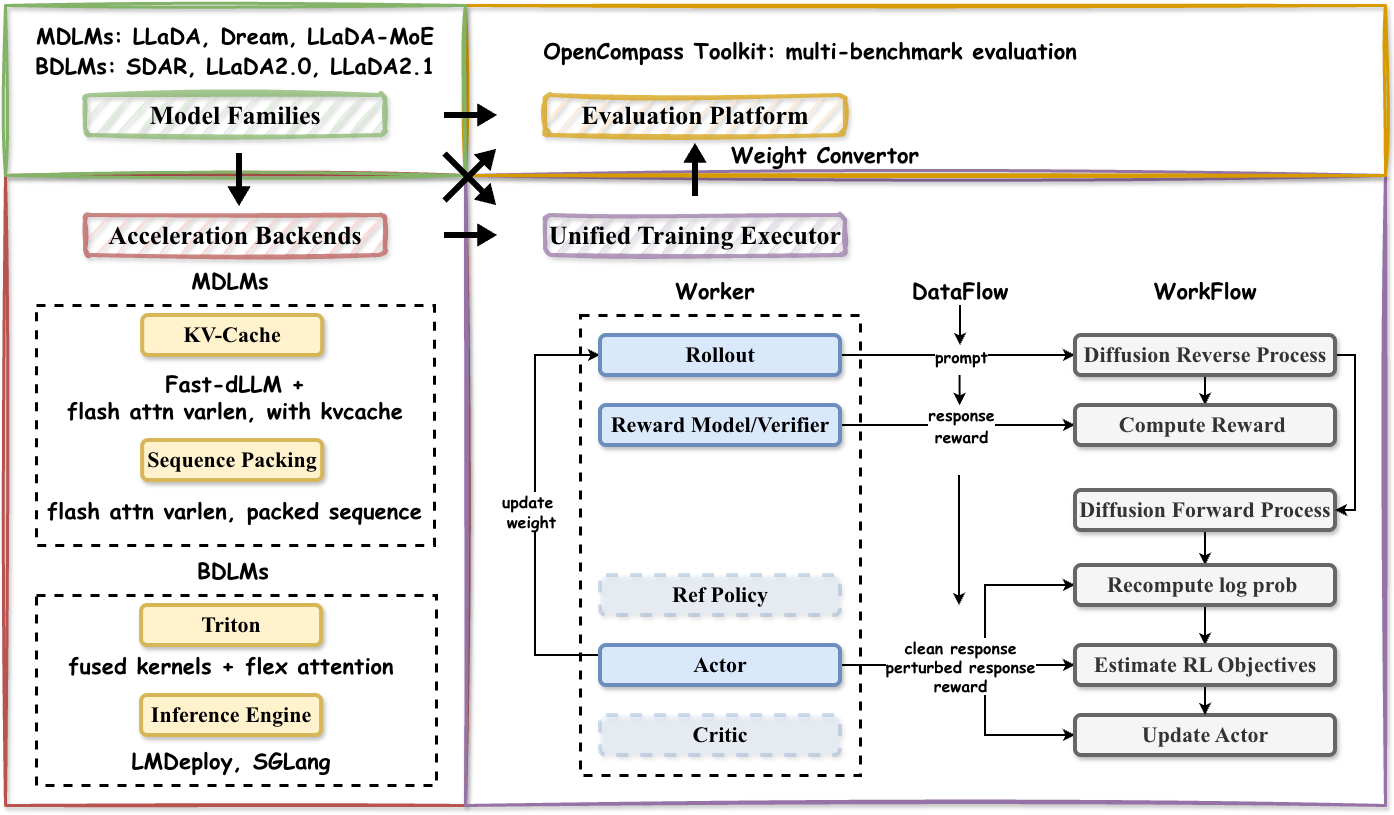}
\vspace{-0.5cm}
\caption{\textbf{High-level view of DARE.} The framework connects dLLM families, model-aware acceleration backends, a unified post-training executor, and an OpenCompass-based evaluation platform under one shared execution stack.}
\label{fig:system-overview}
\end{figure*}

DARE addresses a central systems problem in dLLM research: post-training is still fragmented across model forks, rollout implementations, reward interfaces, and benchmark evaluations. As a result, algorithms are often compared through mismatched execution pipelines rather than within a shared environment.
DARE is designed as a reusable execution layer between open dLLM models, a distributed training backend, and a standardized benchmark stack.
Its design follows three principles: support both masked and block diffusion models under one interface, isolate algorithm-specific logic from shared workflow, and treat evaluation as part of the framework rather than a downstream afterthought, as illustrated in Figure~\ref{fig:system-overview}.

At the implementation level, DARE builds on verl~\cite{sheng2024hybridflow} for distributed training and on OpenCompass~\cite{2023opencompass} for evaluation, while adding dLLM-specific actor, rollout, reward, and model wrappers.
The key design choice is to keep the outer execution skeleton shared and restrict customization to the truly model- or algorithm-specific parts.
In practice, this significantly lowers the engineering cost of integrating upcoming dLLM paradigms/algorithms and changes the practical unit of comparison from 'paper plus custom codebase' to 'algorithm inside a shared executor'

\subsection{Model Families}
\label{subsec:model families}
DARE covers two dominant paradigms in current diffusion language modeling.
On the masked-diffusion side, the framework supports LLaDA~\cite{nie2025large}, Dream~\cite{ye2025dream}, and LLaDA-MoE~\cite{zhu2025llada-moe}, whose training and rollout pipelines revolve around iterative denoising over fully visible sequences.
On the block-diffusion side, DARE supports SDAR~\cite{cheng2025sdar}, LLaDA2.0~\cite{bie2025llada2}, and LLaDA2.1~\cite{bie2026llada2.1}, whose semi-autoregressive structure introduces different rollout order, cache behavior, and attention constraints.
These model families do not merely differ in architecture; they also require different rollout backends, different training optimizations, and in some cases different policy-update paths.

\subsection{Unified Training Executor}
\label{subsec:unified-training-executor}
DARE exposes one post-training executor covering supervised fine-tuning, parameter-efficient fine-tuning, preference optimization, and multiple dLLM-specific reinforcement learning algorithms.

\paragraph{Abstracted as worker, dataflow, workflow.}
The unified executor in Figure~\ref{fig:system-overview} can be understood through three high-level abstractions: worker, dataflow, and workflow.
\emph{\textbf{Workers}} capture the major functional roles in post-training, including rollout, actor, reward model or verifier, and optional reference-policy and critic.
\emph{\textbf{Dataflow}} describes how prompts, responses, reward signals, perturbed trajectories, and log-probability signals move between these workers.
\emph{\textbf{Workflow}} describes the outer optimization loop itself, including rollout (reverse process), reward computation, forward-process when needed, log-probability recomputation, RL-objective estimation, and actor update.
This abstraction also covers SFT, PEFT, and VRPO: their recipes follow the same high-level view, but use a simplified version of the executor with fewer active components and a shorter optimization path than RL.
Under this abstraction, different models, algorithms, and tasks look much more similar than their paper-specific implementations suggest.
The same executor can host multiple dLLM families because what changes is usually a small number of model-aware or algorithm-aware hooks, not the outer training structure.

\paragraph{Shared workflow.}
DARE keeps the high-level PPO-style dataflow shared while exposing diffusion-specific customization only where it is necessary.
At a coarse level, the executor reuses the same skeleton for: \textbf{(i)} rollout generation, \textbf{(ii)} reward computation, \textbf{(iii)} old-policy and optional reference-policy log-probability recomputation, \textbf{(iv)} advantage or return estimation, and \textbf{(v)} actor or critic update.

\begin{table}[t!]
    \vspace{-0.5em}
    \centering
    \caption{\textbf{Post-training methods integrated in DARE across supported dLLM families.} The last column indicates the original implementation or codebase from which the method was adapted, highlighting the fragmented state of prior dLLM post-training infrastructure.}
    \label{tab:integrated-rl-algorithms}
    \resizebox{0.99\columnwidth}{!}{%
    \begin{tabular}{ccccccc||c}
    \toprule
    & \textbf{LLaDA} & \textbf{Dream} & \textbf{SDAR} & \textbf{LLaDA-MoE} & \textbf{LLaDA2.0} & \textbf{LLaDA2.1} & \textbf{\makecell{\textit{Original} \\ \textit{Codebase}}} \\
    \midrule
    \textbf{SFT/PEFT} & \yes & \yes & \yes & \yes & \yes & \yes & model-specific \\
    \textbf{DPO/VRPO}~\cite{zhu2025llada} & \yes & \yes & & \yes & & & closed-source \\
    \textbf{D1}~\cite{zhao2025d1} & \yes & \yes & & \yes & & & D1 \\
    \textbf{Coupled-GRPO}~\cite{gong2025diffucoder} & \yes & \yes & & \yes & & & Open-R1 \\
    \textbf{MDPO}~\cite{he2025mdpo} & \yes & \yes & & \yes & & & Open-R1 \\
    \textbf{CJ-GRPO}~\cite{yang2025taming} & \yes & \yes & & \yes & & & D1 \\
    \textbf{SPG}~\cite{wang2025spg} & \yes & \yes & & \yes & & & D1 \\
    \textbf{BGPO}~\cite{lin2025boundary} & \yes & \yes & \yes & \yes & \yes & \yes & verl \\
    \textbf{EBPO}~\cite{bie2026llada2.1} &  &  &  &  & \yes & \yes & closed-source \\
    \bottomrule
    \end{tabular}
    }
    \vspace{-0.3cm}
    \end{table}

\begin{figure*}[t]
\centering
\includegraphics[width=\textwidth]{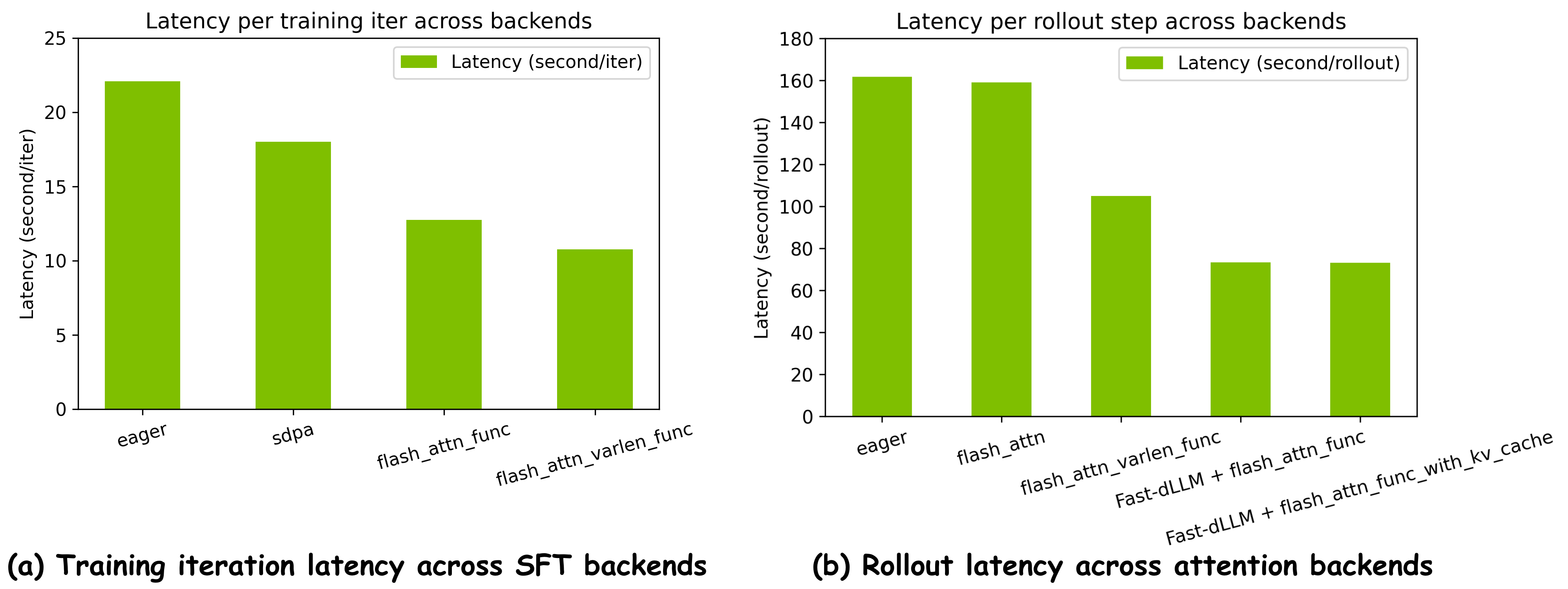}
\vspace{-0.5cm}
\caption{\textbf{Attention backend comparison across training-side and rollout-side for the masked dLLMs pipeline.} (a) shows that optimized attention backends substantially reduce SFT iteration latency relative to eager and sdpa ones. (b) shows that rollout-side backend choice has an even larger effect, with Fast-dLLM-based execution yielding the lowest rollout latency. Together, these results motivate DARE's decoupled optimization of training and rollout paths.}
\label{fig:sft-rollout-latency}
\end{figure*}

\paragraph{Integrated RL Algorithms.}
Table~\ref{tab:integrated-rl-algorithms} summarizes the RL algorithms currently available in the framework.
These include VRPO~\cite{zhu2025llada}, D1~\cite{zhao2025d1}, Coupled-GRPO~\cite{gong2025diffucoder}, MDPO~\cite{he2025mdpo}, CJ-GRPO~\cite{yang2025taming}, SPG~\cite{wang2025spg}, BGPO~\cite{lin2025boundary}, and EBPO~\cite{bie2026llada2.1}.
The last column demonstrates the original implementations of these RL algorithms are scattered across different implementations, while DARE integrates these fragmented implementations into a unified executor.
Most integrated dLLM-RL methods rely on ELBO-based, ELBO-inspired, or trace/trajectory-aware likelihood surrogates. For instance, VRPO, BGPO, SPG, and EBPO represent typical ELBO-based approaches, differing primarily in their approximation of the ELBO bound and whether they employ sequence-level or block-level formulations. In contrast, D1 and Coupled-GRPO draw upon ELBO-inspired objectives, which can be viewed as special cases of ELBO with Monte Carlo sample sizes of 1 and 2, respectively, or equivalently as one-step denoising optimization procedures. Meanwhile, MDPO and CJ-GRPO emphasize step- or trajectory-level optimization.

\paragraph{Algorithm plug-in points.}
What changes across algorithms is not the entire pipeline, but a small number of well-defined hooks.
Depending on the method, these hooks specify the forward corruption process, the trajectory construction rule, the likelihood estimator or bound used in optimization, and the final policy loss. Yet they still share rollout orchestration, reward dispatch, checkpointing, and evaluation.
This separation is what makes side-by-side comparison meaningful: algorithm behavior can be analyzed under matched rollout and verification protocols rather than under entangled implementation differences.

\subsection{Acceleration Backends}
\label{subsec:acceleration-backends}
DARE treats acceleration as a model-aware systems problem rather than as one universal backend choice.
The main reason is that training and rollout exhibit different bottlenecks, and MDLMs (bi-directional attention) and BDLMs (semi-autoregressive) introduce different execution constraints on top of that.

\paragraph{Training-side optimization.}
For supervised fine-tuning and log-probability recomputation, DARE focuses on reducing padding overhead and supporting longer contexts.
For MDLMs such as LLaDA and Dream, the framework uses \texttt{flash\_attn\_varlen\_func} to train efficiently on variable-length batches and combines it with sequence parallelism when scaling to longer contexts.
By skipping unnecessary computation on padding tokens (e.g., \texttt{<PAD>} or \texttt{<EOS>}), this optimization significantly improves training throughput.
As shown in Figure~\ref{fig:sft-rollout-latency}(a), these attention-backend optimizations translate into clear practical gains for MDLM SFT: switching from eager execution to \texttt{flash\_attn\_varlen\_func} reduces per-iteration latency from roughly 22.1 seconds to 10.8 seconds, corresponding to an approximately $2.0\times$ speedup.
For BDLMs such as SDAR, DARE integrates FlexAttention~\cite{dong2024flex} to express semi-autoregressive block constraints while benefiting from compiler-driven kernel optimization.
The framework also supports LoRA~\cite{hu2022lora} for parameter-efficient adaptation.

\paragraph{Rollout-side optimization for MDLMs.}
For masked diffusion models, DARE adopts a key design principle: the best rollout attention backend is not necessarily the best actor update attention backend. Therefore, we decouple the attention backend used during rollout from that used during actor updates.
Rollout benefits from KV-cache reuse and repeated denoising on partially fixed states, so DARE combines Fast-dLLM~\cite{wu2025fast} with \texttt{flash\_attn\_func} or \texttt{flash\_attn\_with\_kvcache} for fast sampling.
Actor update, in contrast, operates on packed sequences with variable valid lengths, where \texttt{flash\_attn\_varlen\_func} is more efficient because it avoids padding-heavy computation.
Figure~\ref{fig:sft-rollout-latency}(b) further shows that rollout-side optimization is even more consequential: Fast-dLLM combined with \texttt{flash\_attn\_func} or \texttt{flash\_attn\_with\_kvcache} reduces rollout latency from about 161.6 seconds under eager execution to about 73.4--73.5 seconds, yielding an approximately $2.2\times$ speedup for the MDLM rollout path.
By explicitly decoupling rollout and training attention backends, DARE achieves an end-to-end RL pipeline speedup of approximately $4\times$ for the MDLM path reported in the repository.

\paragraph{Rollout-side optimization for BDLMs.}
For BDLMs, DARE adopts a different acceleration path that matches their semi-autoregressive structure.
Rollout is accelerated with LMDeploy or SGLang~\cite{2023lmdeploy,zhang2025efficient,sglang2024}, while training uses block-aware attention and model-specific fused operators such as SDAR's logits-free \texttt{fused\_linear\_cross\_entropy}.
The framework also supports online rollout-policy updates to address synchronization issue between rollout and actor.
Together, these design choices provide more than $14\times$ RL pipeline acceleration for the supported BDLM path.


\begin{table*}[t!] 
    \vspace{-0.1cm}
    \centering 
    \caption{\textbf{Baseline performance of LLaDA, Dream, SDAR and LLaDA2.x on various benchmarks.} We apply optimal hyper-parameters for each model under specific benchmarks, such as generation length, block length, denoising steps, etc.}
    \scalebox{0.63}{
    \begin{tabular}{l| cccccc}
    \toprule[1.05pt]
    \textbf{Benchmark/Model} & \multicolumn{1}{c}{\textbf{LLaDA-8B-Instruct}} & \multicolumn{1}{c}{\textbf{Dream-7B-Instruct}} & \multicolumn{1}{c}{\textbf{SDAR-8B-Chat}} & \multicolumn{1}{c}{\textbf{SDAR-30B-A3B}} & \multicolumn{1}{c}{\textbf{LLaDA2.0-mini}} & \multicolumn{1}{c}{\textbf{LLaDA2.1-mini}} \\
    \midrule[1.05pt]
    \multicolumn{7}{c}{\textbf{General QA / Reasoning}} \\
    \midrule[0.8pt]
    \textbf{MMLU} & 65.24 & 66.83 & 77.23 & 79.16 & 72.54 & 69.91 \\
    \textbf{MMLU-Pro} & 36.82 & 31.89 & 56.49 & 25.59 & 57.10 & 57.52 \\
    \textbf{Hellaswag} & 75.30 & 63.23 & 87.59 & 92.81 & 82.35 & 78.00 \\
    \textbf{ARC-C} & 87.80 & 81.36 & 86.78 & 78.98 & 85.76 & 83.39 \\
    \textbf{GPQA} & 31.82 & 26.77 & 41.40 & 36.36 & 34.34 & 34.34 \\
    \midrule[0.8pt]
    \multicolumn{7}{c}{\textbf{Mathematics}} \\
    \midrule[0.8pt]
    \textbf{GSM8k} & 79.68 & 83.24 & 91.36 & 92.49 & 88.48 & 86.13 \\
    \textbf{MATH} & 41.08 & 48.02 & 78.40 & 68.56 & 81.50 & 84.56 \\
    \textbf{AIME24} & 2.08 & 0.83 & 13.33 & 13.33 & 16.67 & 26.67 \\
    \textbf{AIME25} & 0.42 & 0.00 & 16.67 & 6.67 & 23.33 & 26.67 \\
    \textbf{Olympiad} & 9.70 & 12.22 & 24.93 & 32.90 & 38.82 & 40.31 \\
    \midrule[0.8pt]
    \multicolumn{7}{c}{\textbf{Code}} \\
    \midrule[0.8pt]
    \textbf{HumanEval} & 46.34 & 78.05 & 79.88 & 84.15 & 81.10 & 81.10 \\
    \textbf{MBPP} & 38.80 & 56.40 & 71.60 & 52.00 & 64.80 & 62.60 \\
    \bottomrule[1.05pt]
    \end{tabular}
    }
    \label{tab:opencompass-results}
    \vspace{-0.3cm}
    \end{table*}

\begin{table*}[t!]
    \centering
    \caption{\textbf{Mathematics-task results under the DARE.} We report separate algorithm comparisons for LLaDA-8B-Instruct and Dream-7B-Instruct on GSM8K and MATH. Best results are in \textbf{bold}, and second-best results are \underline{underlined}.}
    \vspace{0.1cm}

    \textbf{(a) LLaDA-8B-Instruct}
    \vspace{0.05cm}

    \scalebox{0.8}{
    \begin{tabular}{lccccccc}
    \toprule[1.15pt]
    \textbf{Benchmark/Algorithm} & \textbf{Baseline} & \textbf{d1} & \textbf{Coupled-GRPO} & \textbf{VRPO} & \textbf{CJ-GRPO} & \textbf{SPG} & \textbf{BGPO} \\
    \midrule[1.15pt]
    \textbf{GSM8K} & 76.5 & 83.7 & \underline{85.3} & 81.9 & \textbf{85.6} & 83.5 & 82.3 \\
    \textbf{MATH} & 34.6 & \underline{40.6} & \textbf{41.0} & 35.8 & 39.2 & \underline{40.6} & 40.0 \\
    \bottomrule[1.15pt]
    \end{tabular}
    }

    \vspace{0.25cm}
    \textbf{(b) Dream-7B-Instruct}
    \vspace{0.05cm}

    \scalebox{0.82}{
    \begin{tabular}{lcccccc}
    \toprule[1.15pt]
    \textbf{Benchmark/Algorithm} & \textbf{Baseline} & \textbf{d1} & \textbf{Coupled-GRPO} & \textbf{CJ-GRPO} & \textbf{SPG} & \textbf{BGPO} \\
    \midrule[1.15pt]
    \textbf{GSM8K} & 77.2 & 82.5 & 80.3 & \textbf{85.7} & 59.4 & \underline{83.9} \\
    \textbf{MATH} & 39.6 & \underline{49.7} & 40.4 & \textbf{50.7} & 25.2 & 48.9 \\
    \bottomrule[1.15pt]
    \end{tabular}
    }

    \label{tab:math-results}
    \vspace{-0.25cm}
\end{table*}

\subsection{Evaluation Platform}
\label{subsec:evaluation-platform}

We integrate OpenCompass~\cite{2023opencompass} into DARE to provide a unified and reproducible evaluation platform for dLLMs, and extend it with model-aware execution backends.
For MDLMs, it supports Fast-dLLM-based acceleration~\cite{wu2025fast}; for BDLMs, it supports LMDeploy and SGLang~\cite{2023lmdeploy,sglang2024}.
This avoids the common situation in which post-training code and benchmark code evolve independently and are difficult to reconcile.
DARE covers benchmarks spanning general knowledge, commonsense reasoning, mathematics, olympiad-style reasoning, planning, and code generation, including MMLU~\cite{hendryckstest2021}, MMLU-Pro~\cite{wang2024mmlu}, HellaSwag~\cite{zellers2019hellaswag}, ARC-C~\cite{allenai2018arc}, GSM8K~\cite{cobbe2021training}, MATH~\cite{lightman2023lets}, GPQA~\cite{rein2024gpqa}, AIME2024/2025~\cite{aime2024,aime2025}, OlympiadBench~\cite{he2024olympiadbench}, HumanEval~\cite{chen2021evaluating}, and MBPP~\cite{austin2021program}. Table~\ref{tab:opencompass-results} reports these benchmark results for LLaDA, Dream, SDAR, LLaDA2.0, and LLaDA2.1 reproduced inside DARE.

\section{Empirical Results}
\label{sec:empirical-results}

\begin{table*}[t!]
    \centering
    \caption{\textbf{Code-task results under the DARE.} We report separate algorithm comparisons for LLaDA-8B-Instruct and Dream-7B-Instruct on HumanEval and MBPP. Best results are in \textbf{bold}, and second-best results are \underline{underlined}.}
    \vspace{0.1cm}

    \textbf{(a) LLaDA-8B-Instruct}
    \vspace{0.05cm}

    \scalebox{0.8}{
    \begin{tabular}{lccccccc}
    \toprule[1.15pt]
    \textbf{Benchmark/Algorithm} & \textbf{Baseline} & \textbf{d1} & \textbf{Coupled-GRPO} & \textbf{VRPO} & \textbf{CJ-GRPO} & \textbf{SPG} & \textbf{BGPO} \\
    \midrule[1.15pt]
    \textbf{HumanEval} & 46.9 & 47.6 & 45.1 & \textbf{52.4} & 45.1 & \underline{48.8} & 45.1 \\
    \textbf{MBPP} & 37.9 & 39.1 & 38.1 & \textbf{42.8} & 40.9 & \underline{41.9} & 40.3 \\
    \bottomrule[1.15pt]
    \end{tabular}
    }

    \vspace{0.25cm}
    \textbf{(b) Dream-7B-Instruct}
    \vspace{0.05cm}

    \scalebox{0.82}{
    \begin{tabular}{lcccccc}
    \toprule[1.15pt]
    \textbf{Benchmark/Algorithm} & \textbf{Baseline} & \textbf{d1} & \textbf{Coupled-GRPO} & \textbf{CJ-GRPO} & \textbf{SPG} & \textbf{BGPO} \\
    \midrule[1.15pt]
    \textbf{HumanEval} & 57.9 & \underline{60.7} & \textbf{61.6} & 58.5 & 17.7 & 56.7 \\
    \textbf{MBPP} & 56.2 & 56.5 & \textbf{60.3} & 57.5 & 54.4 & \underline{58.7} \\
    \bottomrule[1.15pt]
    \end{tabular}
    }

    \label{tab:code-results}
    \vspace{-0.25cm}
\end{table*}

\subsection{Reproducibility}
\label{subsec:reproducibility}
\paragraph{Implementation Details.}
We conduct all RL experiments in three tasks, i.e., math, code, and planning~\cite{ye2025dream}, with a unified set of hyper-parameters. Specifically, the rollout group size is set to 8, the block length is 32, and KL regularization is disabled by default. For methods that rely on Monte Carlo likelihood estimation, we set Monte Carlo sampling number to 16.
For mathematical reasoning, we train on the GSM8K training split~\cite{cobbe2021training} together with the MATH  training split~\cite{hendrycks2measuring}. In this task, we use max response length=512, diffusion steps=256, and train for 1 epoch.
For code generation, we use 16K medium-difficulty problems filtered from DeepCoder~\cite{luo2025deepcoder} as the training set, and adopt MBPP and HumanEval as the test sets. 
For planning tasks, we train on Countdown~\cite{tinyzero} and Sudoku~\cite{Arelsudoku} with max response length=256, diffusion steps=128, and 1 epoch, adopting the same test splits as d1~\cite{zhao2025d1}.

\subsection{Main Results}
\label{subsec:main results}
We summarize the main algorithm-comparison results for LLaDA and Dream under the unified DARE executor.

\paragraph{LLaDA as backbone.}
For LLaDA-8B-Instruct, different algorithms dominate in different task regimes.
As math task results in Table~\ref{tab:math-results} (a), CJ-GRPO achieves the best GSM8K result, while Coupled-GRPO is best on MATH, with d1 and SPG remaining competitive.
As code tasks in Table~\ref{tab:code-results} (a), VRPO is strongest on both HumanEval and MBPP, while SPG ranks second.
As planning tasks in Table~\ref{tab:planning-results}, the differences between methods become even sharper, and the ranking changes again: Coupled-GRPO is the strongest on Countdown, whereas BGPO is best on Sudoku.

\paragraph{Dream as backbone.}
For Dream-7B-Instruct, the preferred algorithm is again task-dependent.
As GSM8K and MATH in Table~\ref{tab:math-results} (b), CJ-GRPO is best, with d1 and BGPO also showing strong gains over the baseline.
As code tasks in Table~\ref{tab:code-results} (b), Coupled-GRPO achieves the best results on both HumanEval and MBPP, while d1 and BGPO remain competitive.
At the same time, SPG is much weaker for this backbone on both math and code, suggesting that algorithm robustness can vary substantially across model families.

\textbf{Taken together, the LLaDA and Dream results indicate that there is no single algorithm with uniformly dominant performance across all tasks in our experiments.}
Some methods are particularly effective for math, others for code, and others for planning; moreover, the same method can change rank when the backbone changes.
This is exactly the setting in which a unified executor is most valuable: it makes the absence of a universal winner visible under matched infrastructure.

\subsection{Empirical Findings}
\label{subsec:empirical-findings}

\begin{table*}[t!]
    \centering
    \caption{\textbf{Planning-task results under the DARE.} We report the algorithm comparison for LLaDA-8B-Instruct on Countdown and Sudoku. Best results are in \textbf{bold}, and second-best results are \underline{underlined}.}
    \vspace{0.1cm}

    \scalebox{0.8}{
    \begin{tabular}{lccccccc}
    \toprule[1.15pt]
    \textbf{Benchmark/Algorithm} & \textbf{Baseline} & \textbf{d1} & \textbf{Coupled-GRPO} & \textbf{VRPO} & \textbf{CJ-GRPO} & \textbf{SPG} & \textbf{BGPO} \\
    \midrule[1.15pt]
    \textbf{Countdown} & 16.8 & 10.7 & \textbf{77.9} & 21.5 & \underline{65.2} & 10.1 & 10.0 \\
    \textbf{Sudoku} & 26.2 & \underline{31.8} & 21.3 & 29.0 & 25.0 & 27.9 & \textbf{42.6} \\
    \bottomrule[1.15pt]
    \end{tabular}
    }

    \label{tab:planning-results}
    \vspace{-0.25cm}
\end{table*}

\begin{figure*}[t]
\centering
\includegraphics[width=\textwidth]{figures/training_curve.png}
\vspace{-0.5cm}
\caption{\textbf{Training-curves across different tasks, algorithms and backbones.}}
\label{fig:empirical-findings-curves}
\end{figure*}

Beyond main results, the training curves in Figure~\ref{fig:empirical-findings-curves} may reveal several key findings of the current dLLM RL algorithm and provide some insights.
Across both backbones, d1 and Coupled-GRPO, together with the CJ-GRPO, usually exhibit more stable reward curves (i.e., they are less likely to exhibit abrupt reward collapse).
By contrast, ELBO-based methods may be noticeably more fragile. Their objective estimation accuracy heavily depends on the Monte Carlo sample count, since the ELBO is estimated through diffusion-oriented Monte Carlo surrogates.
When the sample budget is limited, the variance of this estimator is larger, resulting optimization curves can become unstable even when the final objective is competitive on some tasks.

This behavior is visible in several representative runs.
On math with the LLaDA, BGPO shows a clear late-stage collapse.
On Countdown with LLaDA, D1, SPG and BGPO become unstable and eventually degrade sharply.
With the Dream, SPG is also unstable on mathematics and on code, where the reward curve drops substantially instead of converging smoothly.
These cases are not isolated visual artifacts; they align with the broader pattern that ELBO-based optimization is more sensitive to estimator noise and therefore more dependent on sufficient Monte Carlo sample count (for SPG, BGPO) or tighter objective surrogate bound (for BGPO).
The practical significance does not lie in abandoning ELBO based methods. On the contrary, DARE makes system level trade-offs visible: revealing that more precise, stable, and efficient ELBO-objective-based RL algorithms are worth further development.
When this stability evidence is combined with the results in Section~\ref{subsec:main results}, Coupled-GRPO and CJ-GRPO stand out as the most favorable compromises between effectiveness and stability, while d1 remains a robustness and dependable baseline.
Above findings may provide some insights for the development of new dLLM-specific RL algorithms from the community.

\section{Conclusion and Future Work}
\label{sec:conclusion and future work}

We presented DARE, a unified post-training and evaluation framework for diffusion large language models.
Rather than contributing a single new optimization objective, DARE provides a shared execution environment that unifies model families, post-training methods, rollout implementations, reward interfaces, and evaluation pipelines under one reusable stack.
Our experiments show that this framework perspective is useful in practice. DARE supports broad model and benchmark coverage and enables direct comparison of heterogeneous post-training algorithms inside one executor.
There are several natural directions for future work.
On the model side, diffusion (vision-, omni-) language models will require integration.
On the algorithm side, the executor should continue absorbing new estimators, control policies, and stability techniques.
On the systems side, more comprehensive efficiency ablations and deployment-oriented evaluation backends would further strengthen the framework.
We aim to keep DARE aligned with state-of-the-art diffusion large language models and post-training methods, and we welcome contributions and feedback from the open-source community.




\bibliography{colm2026_conference}

@article{nie2025large,
  title={Large language diffusion models},
  author={Nie, Shen and Zhu, Fengqi and You, Zebin and Zhang, Xiaolu and Ou, Jingyang and Hu, Jun and Zhou, Jun and Lin, Yankai and Wen, Ji-Rong and Li, Chongxuan},
  journal={arXiv preprint arXiv:2502.09992},
  year={2025}
}

@article{ye2025dream,
  title={Dream 7B: Diffusion Large Language Models},
  author={Ye, Jiacheng and Xie, Zhihui and Zheng, Lin and Gao, Jiahui and Wu, Zirui and Jiang, Xin and Li, Zhenguo and Kong, Lingpeng},
  journal={arXiv preprint arXiv:2508.15487},
  year={2025}
}

@article{wu2023ar,
  title={Ar-diffusion: Auto-regressive diffusion model for text generation},
  author={Wu, Tong and Fan, Zhihao and Liu, Xiao and Zheng, Hai-Tao and Gong, Yeyun and Jiao, Jian and Li, Juntao and Guo, Jian and Duan, Nan and Chen, Weizhu and others},
  journal={Advances in Neural Information Processing Systems},
  volume={36},
  pages={39957--39974},
  year={2023}
}

@article{austin2021structured,
  title={Structured denoising diffusion models in discrete state-spaces},
  author={Austin, Jacob and Johnson, Daniel D and Ho, Jonathan and Tarlow, Daniel and Van Den Berg, Rianne},
  journal={Advances in neural information processing systems},
  volume={34},
  pages={17981--17993},
  year={2021}
}

@article{sahoo2024simple,
  title={Simple and effective masked diffusion language models},
  author={Sahoo, Subham and Arriola, Marianne and Schiff, Yair and Gokaslan, Aaron and Marroquin, Edgar and Chiu, Justin and Rush, Alexander and Kuleshov, Volodymyr},
  journal={Advances in Neural Information Processing Systems},
  volume={37},
  pages={130136--130184},
  year={2024}
}

@article{lou2023discrete,
  title={Discrete diffusion modeling by estimating the ratios of the data distribution},
  author={Lou, Aaron and Meng, Chenlin and Ermon, Stefano},
  journal={arXiv preprint arXiv:2310.16834},
  year={2023}
}

@article{zheng2024masked,
  title={Masked diffusion models are secretly time-agnostic masked models and exploit inaccurate categorical sampling},
  author={Zheng, Kaiwen and Chen, Yongxin and Mao, Hanzi and Liu, Ming-Yu and Zhu, Jun and Zhang, Qinsheng},
  journal={arXiv preprint arXiv:2409.02908},
  year={2024}
}

@article{gong2024scaling,
  title={Scaling diffusion language models via adaptation from autoregressive models},
  author={Gong, Shansan and Agarwal, Shivam and Zhang, Yizhe and Ye, Jiacheng and Zheng, Lin and Li, Mukai and An, Chenxin and Zhao, Peilin and Bi, Wei and Han, Jiawei and others},
  journal={arXiv preprint arXiv:2410.17891},
  year={2024}
}

@article{ou2024your,
  title={Your absorbing discrete diffusion secretly models the conditional distributions of clean data},
  author={Ou, Jingyang and Nie, Shen and Xue, Kaiwen and Zhu, Fengqi and Sun, Jiacheng and Li, Zhenguo and Li, Chongxuan},
  journal={arXiv preprint arXiv:2406.03736},
  year={2024}
}

@article{nie2024scaling,
  title={Scaling up masked diffusion models on text},
  author={Nie, Shen and Zhu, Fengqi and Du, Chao and Pang, Tianyu and Liu, Qian and Zeng, Guangtao and Lin, Min and Li, Chongxuan},
  journal={arXiv preprint arXiv:2410.18514},
  year={2024}
}

@article{arriola2025block,
  title={Block diffusion: Interpolating between autoregressive and diffusion language models},
  author={Arriola, Marianne and Gokaslan, Aaron and Chiu, Justin T and Yang, Zhihan and Qi, Zhixuan and Han, Jiaqi and Sahoo, Subham Sekhar and Kuleshov, Volodymyr},
  journal={arXiv preprint arXiv:2503.09573},
  year={2025}
}

@article{he2025mdpo,
  title={MDPO: Overcoming the Training-Inference Divide of Masked Diffusion Language Models},
  author={He, Haoyu and Renz, Katrin and Cao, Yong and Geiger, Andreas},
  journal={arXiv preprint arXiv:2508.13148},
  year={2025}
}

@article{zhu2025llada,
  title={LLaDA 1.5: Variance-Reduced Preference Optimization for Large Language Diffusion Models},
  author={Zhu, Fengqi and Wang, Rongzhen and Nie, Shen and Zhang, Xiaolu and Wu, Chunwei and Hu, Jun and Zhou, Jun and Chen, Jianfei and Lin, Yankai and Wen, Ji-Rong and others},
  journal={arXiv preprint arXiv:2505.19223},
  year={2025}
}

@article{zhao2025d1,
  title={d1: Scaling reasoning in diffusion large language models via reinforcement learning},
  author={Zhao, Siyan and Gupta, Devaansh and Zheng, Qinqing and Grover, Aditya},
  journal={arXiv preprint arXiv:2504.12216},
  year={2025}
}

@article{shao2024deepseekmath,
  title={Deepseekmath: Pushing the limits of mathematical reasoning in open language models},
  author={Shao, Zhihong and Wang, Peiyi and Zhu, Qihao and Xu, Runxin and Song, Junxiao and Bi, Xiao and Zhang, Haowei and Zhang, Mingchuan and Li, YK and Wu, Yang and others},
  journal={arXiv preprint arXiv:2402.03300},
  year={2024}
}

@article{guo2025deepseek,
  title={Deepseek-r1: Incentivizing reasoning capability in llms via reinforcement learning},
  author={Guo, Daya and Yang, Dejian and Zhang, Haowei and Song, Junxiao and Zhang, Ruoyu and Xu, Runxin and Zhu, Qihao and Ma, Shirong and Wang, Peiyi and Bi, Xiao and others},
  journal={arXiv preprint arXiv:2501.12948},
  year={2025}
}

@article{cobbe2021training,
  title={Training verifiers to solve math word problems},
  author={Cobbe, Karl and Kosaraju, Vineet and Bavarian, Mohammad and Chen, Mark and Jun, Heewoo and Kaiser, Lukasz and Plappert, Matthias and Tworek, Jerry and Hilton, Jacob and Nakano, Reiichiro and others},
  journal={arXiv preprint arXiv:2110.14168},
  year={2021}
}

@misc{chen2021evaluating,
      title={Evaluating Large Language Models Trained on Code},
      author={Mark Chen and Jerry Tworek and Heewoo Jun and Qiming Yuan and Henrique Ponde de Oliveira Pinto and Jared Kaplan and Harri Edwards and Yuri Burda and Nicholas Joseph and Greg Brockman and Alex Ray and Raul Puri and Gretchen Krueger and Michael Petrov and Heidy Khlaaf and Girish Sastry and Pamela Mishkin and Brooke Chan and Scott Gray and Nick Ryder and Mikhail Pavlov and Alethea Power and Lukasz Kaiser and Mohammad Bavarian and Clemens Winter and Philippe Tillet and Felipe Petroski Such and Dave Cummings and Matthias Plappert and Fotios Chantzis and Elizabeth Barnes and Ariel Herbert-Voss and William Hebgen Guss and Alex Nichol and Alex Paino and Nikolas Tezak and Jie Tang and Igor Babuschkin and Suchir Balaji and Shantanu Jain and William Saunders and Christopher Hesse and Andrew N. Carr and Jan Leike and Josh Achiam and Vedant Misra and Evan Morikawa and Alec Radford and Matthew Knight and Miles Brundage and Mira Murati and Katie Mayer and Peter Welinder and Bob McGrew and Dario Amodei and Sam McCandlish and Ilya Sutskever and Wojciech Zaremba},
      year={2021},
      eprint={2107.03374},
      archivePrefix={arXiv},
      primaryClass={cs.LG}
}

@article{austin2021program,
  title={Program Synthesis with Large Language Models},
  author={Austin, Jacob and Odena, Augustus and Nye, Maxwell and Bosma, Maarten and Michalewski, Henryk and Dohan, David and Jiang, Ellen and Cai, Carrie and Terry, Michael and Le, Quoc and others},
  journal={arXiv preprint arXiv:2108.07732},
  year={2021}
}

@article{khanna2025mercury,
  title={Mercury: Ultra-Fast Language Models Based on Diffusion.},
  author={Khanna, Samar and Kharbanda, Siddhant and Li, Shufan and Varma, Harshit and Wang, Eric and Birnbaum, Sawyer and Luo, Ziyang and Miraoui, Yanis and Palrecha, Akash and Ermon, Stefano and others},
  journal={arXiv preprint arXiv:2506.17298},
  year={2025}
}

@article{song2025seed,
  title={Seed diffusion: A large-scale diffusion language model with high-speed inference},
  author={Song, Yuxuan and Zhang, Zheng and Luo, Cheng and Gao, Pengyang and Xia, Fan and Luo, Hao and Li, Zheng and Yang, Yuehang and Yu, Hongli and Qu, Xingwei and others},
  journal={arXiv preprint arXiv:2508.02193},
  year={2025}
}

@article{geminidiffusion,
  title={Gemini diffusion: our state-of-the-art experimental text diffusion model},
  author={Google DeepMind},
  journal={URL https://deepmind.google/models/gemini-diffusion/},
  year={2025}
}

@article{sheng2024hybridflow,
  title   = {HybridFlow: A Flexible and Efficient RLHF Framework},
  author  = {Guangming Sheng and Chi Zhang and Zilingfeng Ye and Xibin Wu and Wang Zhang and Ru Zhang and Yanghua Peng and Haibin Lin and Chuan Wu},
  year    = {2024},
  journal = {arXiv preprint arXiv: 2409.19256}
}

@misc{2023opencompass,
    title={OpenCompass: A Universal Evaluation Platform for Foundation Models},
    author={OpenCompass Contributors},
    howpublished = {\url{https://github.com/open-compass/opencompass}},
    year={2023}
}

@article{yang2025taming,
  title={Taming Masked Diffusion Language Models via Consistency Trajectory Reinforcement Learning with Fewer Decoding Step},
  author={Yang, Jingyi and Chen, Guanxu and Hu, Xuhao and Shao, Jing},
  journal={arXiv preprint arXiv:2509.23924},
  year={2025}
}

@article{gong2025diffucoder,
  title={DiffuCoder: Understanding and Improving Masked Diffusion Models for Code Generation},
  author={Gong, Shansan and Zhang, Ruixiang and Zheng, Huangjie and Gu, Jiatao and Jaitly, Navdeep and Kong, Lingpeng and Zhang, Yizhe},
  journal={arXiv preprint arXiv:2506.20639},
  year={2025}
}

@article{wang2025spg,
  title={Spg: Sandwiched policy gradient for masked diffusion language models},
  author={Wang, Chenyu and Rashidinejad, Paria and Su, DiJia and Jiang, Song and Wang, Sid and Zhao, Siyan and Zhou, Cai and Shen, Shannon Zejiang and Chen, Feiyu and Jaakkola, Tommi and others},
  journal={arXiv preprint arXiv:2510.09541},
  year={2025}
}

@article{lin2025boundary,
  title={Boundary-guided policy optimization for memory-efficient rl of diffusion large language models},
  author={Lin, Nianyi and Zhang, Jiajie and Hou, Lei and Li, Juanzi},
  journal={arXiv preprint arXiv:2510.11683},
  year={2025}
}

@article{fathi2025unifying,
  title={Unifying autoregressive and diffusion-based sequence generation},
  author={Fathi, Nima and Scholak, Torsten and No{\"e}l, Pierre-Andr{\'e}},
  journal={arXiv preprint arXiv:2504.06416},
  year={2025}
}

@article{han2022ssd,
  title={Ssd-lm: Semi-autoregressive simplex-based diffusion language model for text generation and modular control},
  author={Han, Xiaochuang and Kumar, Sachin and Tsvetkov, Yulia},
  journal={arXiv preprint arXiv:2210.17432},
  year={2022}
}

@article{cheng2025sdar,
  title={Sdar: A synergistic diffusion-autoregression paradigm for scalable sequence generation},
  author={Cheng, Shuang and Bian, Yihan and Liu, Dawei and Zhang, Linfeng and Yao, Qian and Tian, Zhongbo and Wang, Wenhai and Guo, Qipeng and Chen, Kai and Qi, Biqing and others},
  journal={arXiv preprint arXiv:2510.06303},
  year={2025}
}

@article{wu2025fast,
  title={Fast-dllm: Training-free acceleration of diffusion llm by enabling kv cache and parallel decoding},
  author={Wu, Chengyue and Zhang, Hao and Xue, Shuchen and Liu, Zhijian and Diao, Shizhe and Zhu, Ligeng and Luo, Ping and Han, Song and Xie, Enze},
  journal={arXiv preprint arXiv:2505.22618},
  year={2025}
}

@misc{2023lmdeploy,
    title={LMDeploy: A Toolkit for Compressing, Deploying, and Serving LLM},
    author={LMDeploy Contributors},
    howpublished = {\url{https://github.com/InternLM/lmdeploy}},
    year={2023}
}

@article{zhang2025efficient,
  title={Efficient Mixed-Precision Large Language Model Inference with TurboMind},
  author={Zhang, Li and Jiang, Youhe and He, Guoliang and Chen, Xin and Lv, Han and Yao, Qian and Fu, Fangcheng and Chen, Kai},
  journal={arXiv preprint arXiv:2508.15601},
  year={2025}
}

@misc{sglang2024,
    title={SGLang},
    author={SGLang Contributors},
    howpublished = {\url{https://github.com/sgl-project/sglang}},
    year={2024}
}

@article{dao2022flashattention,
  title={Flashattention: Fast and memory-efficient exact attention with io-awareness},
  author={Dao, Tri and Fu, Dan and Ermon, Stefano and Rudra, Atri and R{\'e}, Christopher},
  journal={Advances in neural information processing systems},
  volume={35},
  pages={16344--16359},
  year={2022}
}

@article{dao2023flashattention,
  title={Flashattention-2: Faster attention with better parallelism and work partitioning},
  author={Dao, Tri},
  journal={arXiv preprint arXiv:2307.08691},
  year={2023}
}

@article{hendryckstest2021,
    title={Measuring Massive Multitask Language Understanding},
    author={Dan Hendrycks and Collin Burns and Steven Basart and Andy Zou and Mantas Mazeika and Dawn Song and Jacob Steinhardt},
    journal={Proceedings of the International Conference on Learning Representations (ICLR)},
    year={2021}
}

@article{wang2024mmlu,
  title={Mmlu-pro: A more robust and challenging multi-task language understanding benchmark},
  author={Wang, Yubo and Ma, Xueguang and Zhang, Ge and Ni, Yuansheng and Chandra, Abhranil and Guo, Shiguang and Ren, Weiming and Arulraj, Aaran and He, Xuan and Jiang, Ziyan and others},
  journal={Advances in Neural Information Processing Systems},
  volume={37},
  pages={95266--95290},
  year={2024}
}

@inproceedings{zellers2019hellaswag,
    title={HellaSwag: Can a Machine Really Finish Your Sentence?},
    author={Zellers, Rowan and Holtzman, Ari and Bisk, Yonatan and Farhadi, Ali and Choi, Yejin},
    booktitle ={Proceedings of the 57th Annual Meeting of the Association for Computational Linguistics},
    year={2019}
}

@article{allenai2018arc,
      author={Peter Clark  and Isaac Cowhey and Oren Etzioni and Tushar Khot and Ashish Sabharwal and Carissa Schoenick and Oyvind Tafjord},
      title={Think you have Solved Question Answering? Try ARC, the AI2 Reasoning Challenge},
      journal={arXiv:1803.05457v1},
      year={2018},
}

@article{lightman2023lets,
      title={Let's Verify Step by Step}, 
      author={Lightman, Hunter and Kosaraju, Vineet and Burda, Yura and Edwards, Harri and Baker, Bowen and Lee, Teddy and Leike, Jan and Schulman, John and Sutskever, Ilya and Cobbe, Karl},
      journal={arXiv preprint arXiv:2305.20050},
      year={2023}
}

@inproceedings{rein2024gpqa,
  title={Gpqa: A graduate-level google-proof q\&a benchmark},
  author={Rein, David and Hou, Betty Li and Stickland, Asa Cooper and Petty, Jackson and Pang, Richard Yuanzhe and Dirani, Julien and Michael, Julian and Bowman, Samuel R},
  booktitle={First Conference on Language Modeling},
  year={2024}
}

@misc{aime2024,
    title={AIME2024},
    author={AIME 2024 Contributors},
    howpublished = {\url{https://huggingface.co/datasets/Maxwell-Jia/AIME_2024}},
    year={2024}
}

@misc{aime2025,
    title={AIME2025},
    author={AIME 2025 Contributors},
    howpublished = {\url{https://huggingface.co/datasets/math-ai/aime25}},
    year={2025}
}

@article{he2024olympiadbench,
  title={Olympiadbench: A challenging benchmark for promoting agi with olympiad-level bilingual multimodal scientific problems},
  author={He, Chaoqun and Luo, Renjie and Bai, Yuzhuo and Hu, Shengding and Thai, Zhen Leng and Shen, Junhao and Hu, Jinyi and Han, Xu and Huang, Yujie and Zhang, Yuxiang and others},
  journal={arXiv preprint arXiv:2402.14008},
  year={2024}
}

@article{dong2024flex,
  title={Flex attention: A programming model for generating optimized attention kernels},
  author={Dong, Juechu and Feng, Boyuan and Guessous, Driss and Liang, Yanbo and He, Horace},
  journal={arXiv preprint arXiv:2412.05496},
  year={2024}
}

@article{hu2022lora,
  title={Lora: Low-rank adaptation of large language models.},
  author={Hu, Edward J and Shen, Yelong and Wallis, Phillip and Allen-Zhu, Zeyuan and Li, Yuanzhi and Wang, Shean and Wang, Lu and Chen, Weizhu and others},
  journal={ICLR},
  volume={1},
  number={2},
  pages={3},
  year={2022}
}

@article{bie2025llada2,
  title={Llada2. 0: Scaling up diffusion language models to 100b},
  author={Bie, Tiwei and Cao, Maosong and Chen, Kun and Du, Lun and Gong, Mingliang and Gong, Zhuochen and Gu, Yanmei and Hu, Jiaqi and Huang, Zenan and Lan, Zhenzhong and others},
  journal={arXiv preprint arXiv:2512.15745},
  year={2025}
}

@article{bie2026llada2.1,
  title={LLaDA2. 1: Speeding Up Text Diffusion via Token Editing},
  author={Bie, Tiwei and Cao, Maosong and Cao, Xiang and Chen, Bingsen and Chen, Fuyuan and Chen, Kun and Du, Lun and Feng, Daozhuo and Feng, Haibo and Gong, Mingliang and others},
  journal={arXiv preprint arXiv:2602.08676},
  year={2026}
}

@article{yang2026rho,
  title={$\rho$-$\texttt{EOS}$: Training-free Bidirectional Variable-Length Control for Masked Diffusion LLMs},
  author={Yang, Jingyi and Jiang, Yuxian and Shao, Jing},
  journal={arXiv preprint arXiv:2601.22527},
  year={2026}
}

@article{wang2025revolutionizing,
  title={Revolutionizing reinforcement learning framework for diffusion large language models},
  author={Wang, Yinjie and Yang, Ling and Li, Bowen and Tian, Ye and Shen, Ke and Wang, Mengdi},
  journal={arXiv preprint arXiv:2509.06949},
  year={2025}
}

@article{zhu2025llada-moe,
  title={Llada-moe: A sparse moe diffusion language model},
  author={Zhu, Fengqi and You, Zebin and Xing, Yipeng and Huang, Zenan and Liu, Lin and Zhuang, Yihong and Lu, Guoshan and Wang, Kangyu and Wang, Xudong and Wei, Lanning and others},
  journal={arXiv preprint arXiv:2509.24389},
  year={2025}
}

@article{li2025beyond,
  title={Beyond fixed: Training-free variable-length denoising for diffusion large language models},
  author={Li, Jinsong and Dong, Xiaoyi and Zang, Yuhang and Cao, Yuhang and Wang, Jiaqi and Lin, Dahua},
  journal={arXiv preprint arXiv:2508.00819},
  year={2025}
}

@article{zhu2025dirl,
  title={Dirl: An efficient post-training framework for diffusion language models},
  author={Zhu, Ying and Wan, Jiaxin and Liu, Xiaoran and He, Siyang and Wang, Qiqi and Guo, Xu and Liang, Tianyi and Huang, Zengfeng and He, Ziwei and Qiu, Xipeng},
  journal={arXiv preprint arXiv:2512.22234},
  year={2025}
}

@article{hendrycks2measuring,
  title={Measuring Mathematical Problem Solving With the MATH Dataset},
  author={Hendrycks, Dan and Burns, Collin and Kadavath, Saurav and Arora, Akul and Basart, Steven and Tang, Eric and Song, Dawn and Steinhardt, Jacob},
  journal={Sort},
  volume={2},
  number={4},
  pages={0--6}
}

@article{luo2025deepcoder,
  title={Deepcoder: A fully open-source 14b coder at o3-mini level},
  author={Luo, Michael and Tan, Sijun and Huang, Roy and Patel, Ameen and Ariyak, Alpay and Wu, Qingyang and Shi, Xiaoxiang and Xin, Rachel and Cai, Colin and Weber, Maurice and others},
  journal={Notion Blog},
  volume={1},
  year={2025}
}

@misc{tinyzero,
author       = {Jiayi Pan and Junjie Zhang and Xingyao Wang and Lifan Yuan and Hao Peng and Alane Suhr},
title        = {TinyZero},
howpublished = {https://github.com/Jiayi-Pan/TinyZero},
note         = {Accessed: 2025-01-24},
year         = {2025}
}

@misc{Arelsudoku,
author       = {Arel},
title        = {Arel’s sudoku generator},
howpublished = {https://www.ocf.berkeley.edu/~arel/sudoku/main.html},
year         = {2025}
}

@article{zhao2025diffpo,
  title={Diffpo: Training diffusion llms to reason fast and furious via reinforcement learning},
  author={Zhao, Hanyang and Liang, Dawen and Tang, Wenpin and Yao, David and Kallus, Nathan},
  journal={arXiv preprint arXiv:2510.02212},
  year={2025}
}

@article{chen2025conditional,
  title={Conditional Advantage Estimation for Reinforcement Learning in Large Reasoning Models},
  author={Chen, Guanxu and Li, Yafu and Jiang, Yuxian and Qian, Chen and Ren, Qihan and Yang, Jingyi and Cheng, Yu and Liu, Dongrui and Shao, Jing},
  journal={arXiv preprint arXiv:2509.23962},
  year={2025}
}
\bibliographystyle{colm2026_conference}


\end{document}